# Nucleus subtype classification using inter-modality learning

Lucas W. Remedios[a*], Shunxing Bao[b], Samuel W. Remedios[c,d], Ho Hin Lee[a], Leon Y. Cai[e], Thomas Li[e], Ruining Deng[a], Can Cui[a], Jia Li[f], Qi Liu[f,g], Ken S. Lau[g,h,i], Joseph T. Roland[h], Mary K. Washington[n], Lori A. Coburn[j,k,l,m], Keith T. Wilson[j,k,l,m,n], Yuankai Huo[a,b], Bennett A. Landman[a,b,e]

[a]Vanderbilt University, Department of Computer Science, Nashville, USA; [b]Vanderbilt University, Department of Electrical and Computer Engineering, Nashville, USA; [c]Johns Hopkins University, Department of Computer Science, Baltimore, USA; [d]National Institutes of Health, Department of Radiology and Imaging Sciences, Bethesda, USA; [e]Vanderbilt University, Department of Biomedical Engineering, Nashville, USA; [f]Vanderbilt University Medical Center, Department of Biostatistics, Nashville, USA; [g]Vanderbilt University Medical Center, Center for Quantitative Sciences, Nashville, USA; [h]Vanderbilt University Medical Center, Epithelial Biology Center, Nashville, USA; [i]Vanderbilt University School of Medicine, Department of Cell and Developmental Biology, Nashville, USA; [j]Vanderbilt University Medical Center, Division of Gastroenterology, Hepatology, and Nutrition, Department of Medicine, Nashville, USA; [k]Vanderbilt University Medical Center, Vanderbilt Center for Mucosal Inflammation and Cancer, Nashville, USA; [l]Vanderbilt University School of Medicine, Program in Cancer Biology, Nashville, USA; [m]Veterans Affairs Tennessee Valley Healthcare System, Nashville, TN, USA; [n]Vanderbilt University Medical Center, Department of Pathology, Microbiology, and Immunology, Nashville, TN, USA

## ABSTRACT

Understanding the way cells communicate, co-locate, and interrelate is essential to understanding human physiology. Hematoxylin and eosin (H&E) staining is ubiquitously available both for clinical studies and research. The Colon Nucleus Identification and Classification (CoNIC) Challenge has recently innovated on robust artificial intelligence labeling of six cell types on H&E stains of the colon. However, this is a very small fraction of the number of potential cell classification types. Specifically, the CoNIC Challenge is unable to classify epithelial subtypes (progenitor, endocrine, goblet), lymphocyte subtypes (B, helper T, cytotoxic T), or connective subtypes (fibroblasts, stromal). In this paper, we propose to use inter-modality learning to label previously un-labelable cell types on virtual H&E. We leveraged multiplexed immunofluorescence (MxIF) histology imaging to identify 14 subclasses of cell types. We performed style transfer to synthesize virtual H&E from MxIF and transferred the higher density labels from MxIF to these virtual H&E images. We then evaluated the efficacy of learning in this approach. We identified helper T and progenitor nuclei with positive predictive values of $0.34 \pm 0.15$ (prevalence $0.03 \pm 0.01$) and $0.47 \pm 0.1$ (prevalence $0.07 \pm 0.02$) respectively on virtual H&E. This approach represents a promising step towards automating annotation in digital pathology.

## 1. INTRODUCTION

Hematoxylin and eosin (H&E) stains are ubiquitous in pathology[1], coloring nuclei blue and the cytoplasm and other parts of the tissue pink[2]. Accurate manual annotation of fine-grained anatomical structures in H&E is challenging for non-pathologists[3], which makes large scale manual annotation of subtle structures time intensive. In 2022, the CoNIC Challenge released a dataset of colon H&E with nucleus types[3,4] annotated using a complicated and repetitive approach that involved both automatic nucleus annotation and refinement based on feedback from trained pathologists[3,4]. The development of precise nucleus classification algorithms on H&E is important for the study of disease on this routinely collected stain type.

In contrast to H&E, multiplexed immunofluorescence (MxIF) enables staining and imaging of the same tissue many times via bleaching and re-staining[5]. MxIF multi-channel imaging provides more rich information about tissue structure than H&E, as each stain binds to a specific subset of the tissue. Ideally, MxIF information could be translated to H&E without the need to physically perform the MxIF staining procedure. Following this concept, Nadarajan et al. performed semantic segmentation of simple structures on H&E using MxIF-derived labels from the same tissue[6]. Their approach required paired H&E and MxIF stains, which is not always available. In a follow-up paper, the same group used a conditional generative adversarial network (GAN) to create virtual H&E from MxIF[2]. They then trained a semantic segmentation model on the virtual H&E with MxIF-derived labels to semantically segment 4 structures (all nuclei, cytoplasm, membranes, background) and evaluated on H&E. Further work has been conducted in this area with Han et al. having designed a model that learned to classify 4 types of cells (ER+, PR+, HER2+, and Ki67+) from real H&E by leveraging paired MxIF information[7].

We take these previous works a step further by investigating the task of classifying 14 types of nuclei on virtual H&E with the assistance of MxIF (Figure 1). The contribution of this work is to enable the first approach to automatically identify helper T and progenitor nuclei on virtual H&E.

## 2. METHODS

We studied an in-house dataset of MxIF. These images were style transferred to the H&E domain (§2.3). To distinguish acquired H&E and synthesized H&E, henceforth, we refer to real H&E and virtual H&E, respectively. We used deep learning to learn to classify 14 types of nuclei on virtual H&E in a supervised multi-class classification approach (§2.4). The label information was derived from the MxIF. Specifically, labels were generated for nucleus classes via combinations of 17 out of 27 MxIF stains (§2.2 and Table 1). The CycleGAN-synthesized virtual H&E was generated using all 27 MxIF stains. We trained a ResNet[8] on virtual H&E with MxIF-derived nucleus class labels and evaluated the models on withheld virtual H&E. An overview of our methods can be seen in Figure 2.

### 2.1 In-house MxIF Data

Samples were studied in deidentified form from Vanderbilt University Medical Center under Institutional Review Board approval (IRB #191738 and #191777). The imaged samples depicted either normal tissue, inactive Crohn's disease, or active Crohn's disease and were formalin-fixed and paraffin-embedded. We used 28 whole slides imaged at 0.32 microns per pixel (14 from the ascending colon and 14 from the terminal ileum). These slides came from 20 patients. Most patients had 1 slide, but could have up to 4 slides. We studied 17 out of 27 stain channels to annotate cell types on MxIF: NaKATPase, PanCK, Muc2, CgA, Vimentin, DAPI, SMA, Sox9, OLFM4, Lysozyme, CD45, CD20, CD68, CD11B, CD3d, CD8, and CD4 (see Table 1 for reasoning). Although we used 17 stains for annotation, we annotated 14 cell types. This is because each of the 17 stains did not always directly map to a unique cell type. These stains came from a subset of the staining protocol described in a previous work from our group[9].

### 2.2 Label Generation

To determine positive+ and negative- nuclei for each stain, we applied stain-wise thresholds that were manually selected by a senior digital pathology researcher. These thresholds were determined separately by stain channel for each MxIF image. To determine the location of each nucleus, we performed inference with the DeepCell Mesmer model[10] on the MxIF DAPI and Muc2 channels. Each nucleus was then categorized as positive+ or negative- for each stain type by computing its mean stain intensity and applying the manual thresholds.

We assigned each nucleus a single class label based on a series of biological rules (Table 1). After label generation, we had 14 nucleus classes: goblet, endocrine, epithelial, progenitor, fibroblast, stromal, monocyte, macrophage, helper T, cytotoxic T, T cell receptor, B, myeloid, and leukocyte.

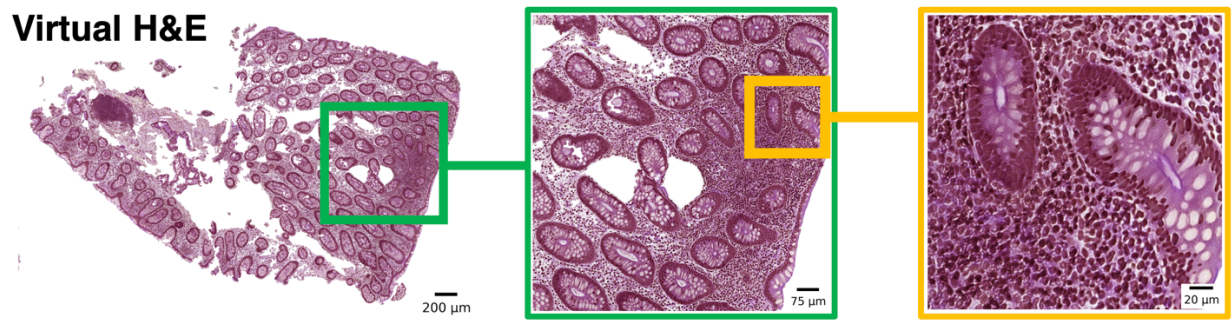

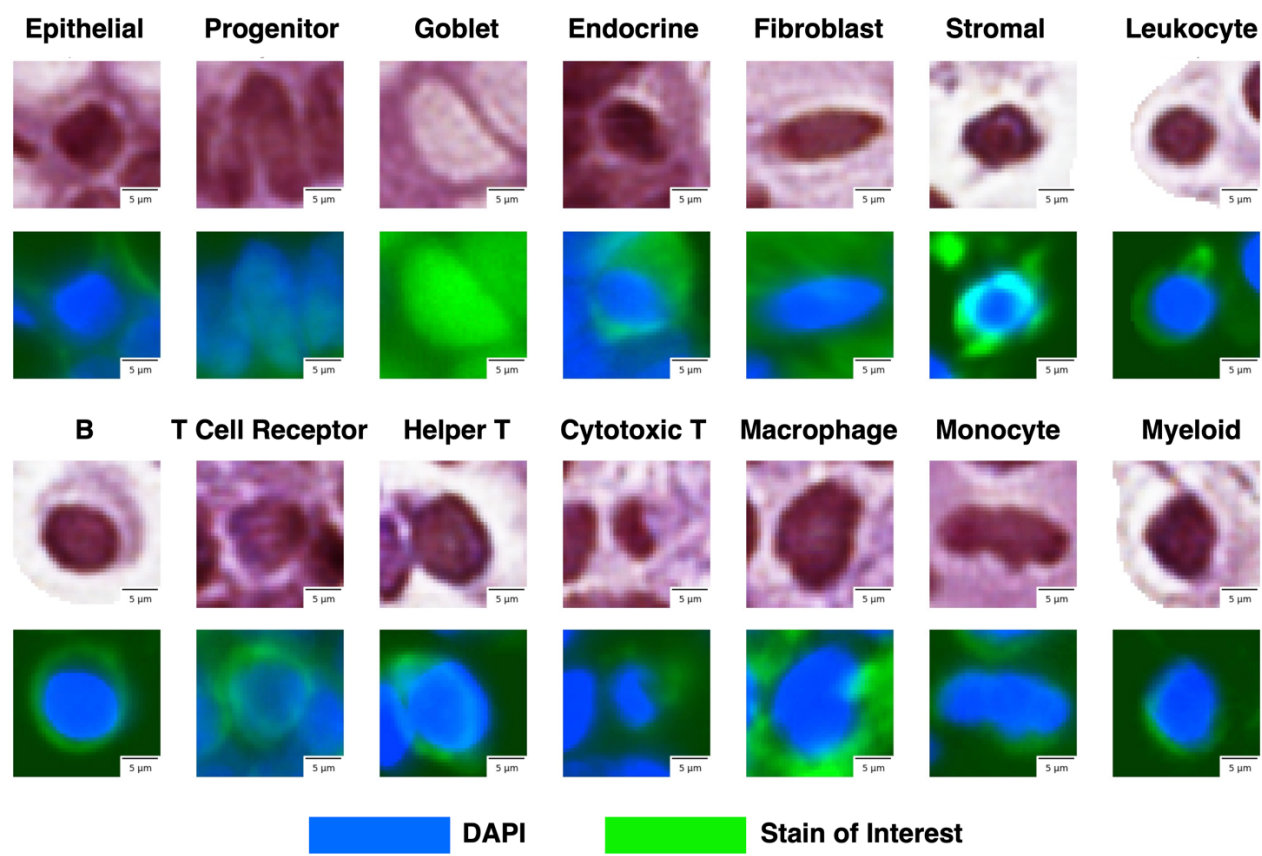

Figure 1. We leveraged inter-modality learning to investigate identification of cells on virtual H&E staining that are traditionally viewed with specialized staining. The realistic quality of our virtual H&E holds at multiple scales (top row). Representative nuclei from each of our 14 classes in both virtual H&E and MxIF illustrate intensity and morphological variation across cell types (lower section). Green is used to denote the MxIF stain of interest, which is a different stain for each of the 14 cell types in this figure. While the signal to identify these classes of nuclei is present in MxIF, the nucleus classes are more difficult to distinguish on virtual H&E.

Table 1. A sequential decision process was used for generating labels for the 14 nucleus classes from MxIF. The bottom rows show final annotations for the 14 classes.

| Step | Purpose | Stain Combinations |
|---|---|---|
| 1 | Group Epi+ nuclei | NaKATPase+ or PanCK+ or Muc2+ or CgA+ |
| 2 | Group Stroma+ nuclei | Vimentin+ or SMA+ |
| 3 | Exclude nuclei that are both Epi+ and Stroma+ | Exclude nuclei that are marked as (Epi+ and Stroma+) |
| 4 | Group Immune+ nuclei | Group nuclei that are CD45+ or CD20+ or CD68+ or CD11B+ or Lysozyme+ or CD3d+ or CD8+ or CD4+ |
| 5 | Remove immune conflicts for macrophage nuclei across all nuclei | Exclude nuclei where (CD68+ and CD3d+), (CD68+ and CD20+), (CD68+ and CD4+), (CD68+ and CD8+), or (CD68+ and CD11B+) |
| 6 | Remove immune conflicts for monocyte nuclei across all nuclei | Exclude nuclei where (CD11B+ and CD3d+), (CD11B+ and CD20+), (CD11B+ and CD4+), (CD11B+ and CD8+), or (CD11B+ and CD68+) |
| 7 | Remove immune conflicts for B cell nuclei across all nuclei | Exclude nuclei where (CD20+ and CD3d+), (CD20+ and CD4+), or (CD20+ and CD8+) |
| 8 | Remove conflicts for helper T and cytotoxic T nuclei across all nuclei | Exclude nuclei where (CD3d- and CD45- and CD4+), (CD3d- and CD45- and CD8+), or (CD4+ and CD8+) |
| 9 | Group Progenitor+ nuclei | Sox9+ or OLFM4+ |
| 10 | Exclude nuclei that are not in either the epithelium or stroma | Exclude nuclei that are (Epi- and Stroma-) |
| 11 | Remove conflicts for goblet cells across all nuclei | Exclude nuclei where (Muc2+ and Immune+), (Muc2+ and Progenitor+), or (Muc2+ and SMA+) |
| 12 | Remove conflicts for endocrine nuclei across all nuclei | Exclude nuclei where (CgA+ and Immune+), (CgA+ and SMA+), (CgA+ and Progenitor+), or (CgA+ and Muc2+) |
| 13 | Remove conflicts for fibroblasts across all nuclei | Exclude nuclei that are (SMA+ and Immune+) |
| 14 | Remove conflicts for progenitors across all nuclei | Exclude nuclei where (Immune+ and Progenitor+) |
| 15 | Remove nuclei that are negative for all the large groupings | Exclude nuclei where (Epi- and Stroma- and Progenitor- and Immune-) |
| 16 | Remove any immune nuclei from Epi+ group | Exclude Epi+ nuclei where (Epi+ and Immune+) |
| 17 | Final annotation for goblet cells | Group Epi+ nuclei where (Muc2+ and Progenitor-) |
| 18 | Final annotation for endocrine cells | Group Epi+ nuclei where (CgA+ and Progenitor-) |
| 19 | Final annotation for epithelial cells | Group Epi+ nuclei where (CgA- and Progenitor- and Muc2-) |
| 20 | Group stromal/fibroblasts | Group nuclei that are Stroma+ and Immune- |
| 21 | Final annotation for fibroblasts | Group stromal/fibroblast nuclei where (SMA+ and Progenitor-) |
| 22 | Final annotation for stromal nuclei | Group stromal/fibroblast nuclei where (SMA- and Progenitor-) |
| 23 | Final annotation for myeloid nuclei | Group Immune+ nuclei that are ((Lysozyme+ and CD68- and CD11B- and Progenitor- and CD20-) and CD3d- and CD8- and CD4-)) |
| 24 | Final annotation for helper T nuclei | Group Immune+ nuclei where (CD4+ and Progenitor-) |
| 25 | Final annotation for cytotoxic T nuclei | Group Immune+ nuclei where (CD8+ and Progenitor-) |
| 26 | Final annotation for T cell receptors | Group Immune+ nuclei where (CD3d+ and CD4- and CD8-) |
| 27 | Final annotation for monocyte nuclei | Group Immune+ nuclei where (CD11b+ and CD3d- and Progenitor- and CD4- and CD8-) |
| 28 | Final annotation for macrophage nuclei | Group Immune+ nuclei where (CD68+ and CD3d- and Progenitor- and CD4- and CD8-) |
| 29 | Final annotation for B cell nuclei | Group Immune+ nuclei where (CD20+ and CD68- and CD3d- and Progenitor- and CD4- and CD8-) |
| 30 | Final annotation for leukocyte nuclei | Group Immune+ nuclei where (CD45+ and CD20- and CD68- and CD3d- and Progenitor- and CD4- and CD8- and CD11B- and Lysozyme-) |
| 31 | Final annotation for progenitor nuclei | Group all nuclei that are Progenitor+ |

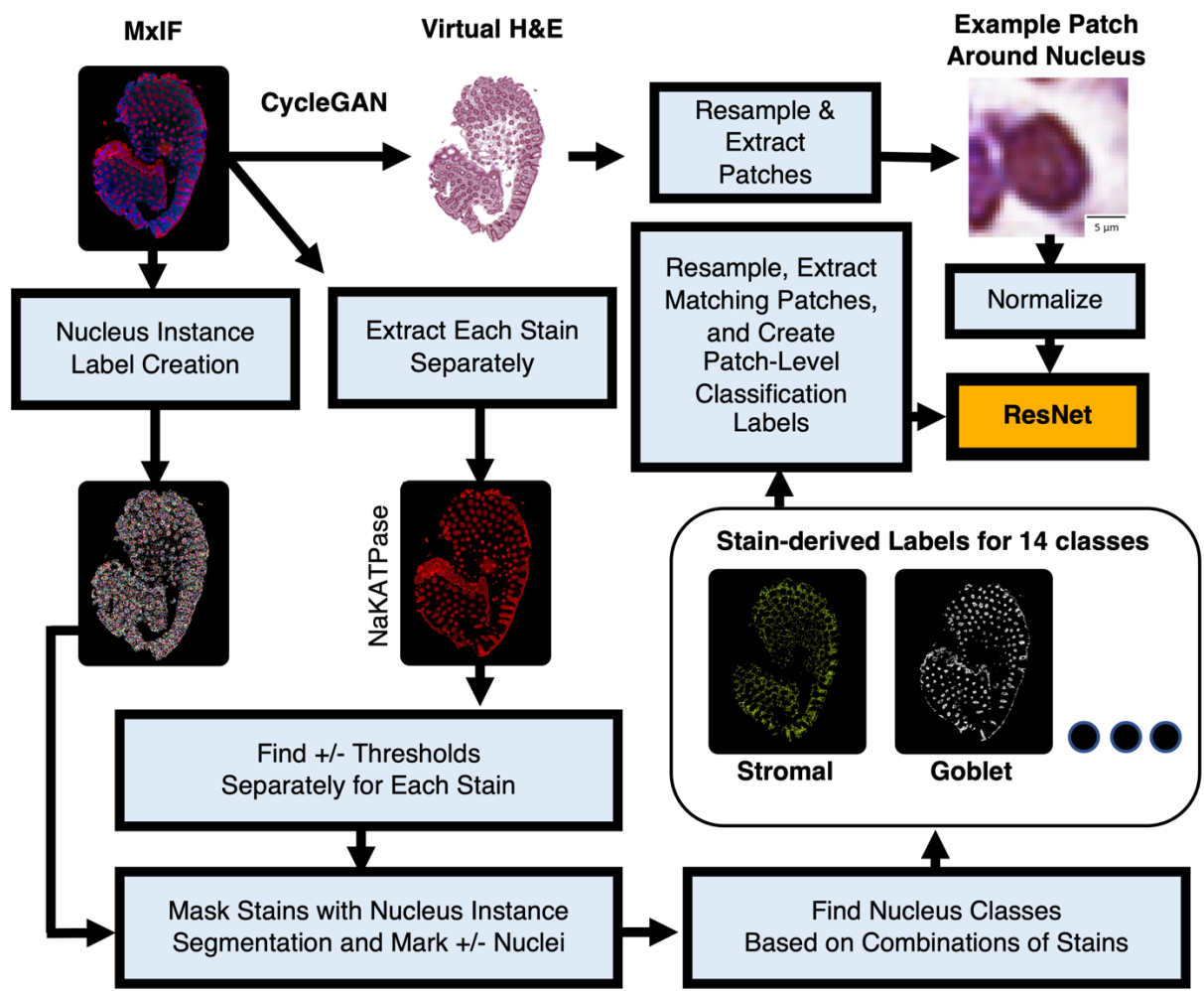

Figure 2. Our approach is similar to Han et al.,[7] with the additions of learning nucleus classification from virtual data and targeting 14 classes rather than 4.

**2.3 Virtual H&E via Style Transfer**

To train the models for classification of nuclei on virtual H&E, we needed image-label pairs of virtual H&E (from MxIF) and cell labels (from MxIF). The virtual H&E was inferred from all 27 MxIF stains using a pretrained model that performed style transfer. This model was trained on in-house MxIF and in-house real H&E. The architecture and training strategy (named "Proposed-(8)") were described in detail in a previous work from our team[9].

**2.4 Classification & Cross-validation**

We selected ResNet-18[8] as our nucleus classification model. The model was initialized from the PyTorch default ResNet pretrained on ImageNet. We replaced each batchnorm layer with the PyTorch 2D instance norm layer using default parameters in order to remove effects from difference of batch size and balance between training and testing. This model was trained for classification on image patches of virtual H&E. The patches were resampled to a standard H&E resolution of 0.5 microns per pixel. Our patch size was 41×41 pixels. Each patch was taken with a nucleus in the center and normalized between 0 and 1. A single class label was assigned to each patch, corresponding to the center nucleus. Visual examples of this patch size and approach can be seen in the subplots in Figure 1. We uniformly pulled from the 14 classes for each example in each batch to handle class imbalance. The model was trained for 20,000 steps using a batch size of 256, Adam optimizer, learning rate of 1e-3, crossentropy loss, and one cycle learning rate scheduler.

We performed five-fold cross-validation. To avoid data contamination, training, validation, and testing data were split at the patient level. To maintain consistency across folds, we specified that in each fold, the training data contained 12 patients, validation contained 4 patients, and testing contained 4 patients. To reduce bias, we always included data from both the ascending colon and terminal ileum, healthy and diseased, in each training, validation, and testing set. We then selected weights for evaluation based on the step with the lowest validation loss for each fold. We evaluated the nucleus classification models on the corresponding virtual H&E testing data.

All of the classification code was implemented in Python 3.8 using PyTorch 1.12.1 and Torchvision 0.13.1. Additionally, all training and inference was performed using an Nvidia RTX A6000.

## 3. RESULTS

The classification accuracy of the ResNet showed learnability for a subset of classes (Figure 3). These classes were helper T, macrophage, epithelial, progenitor, endocrine, goblet, and fibroblast nuclei. Looking in more detail at classification performance, we computed the positive predictive value (PPV), negative predictive value (NPV), and prevalence (Figure 4). When prevalence is low, we expect PPV to be low and NPV to be high. Likewise, when prevalence is high, we expect PPV to be high and NPV to be low. A cutoff for reliable classification could be selected based on PPV. Figure 4 illustrates an example PPV cutoff of 0.3 for reliable learning. The classes above this threshold are helper T, epithelial, progenitor, goblet, fibroblast, and stromal nuclei. In detail, the PPV, NPV, and prevalence for each class are given respectively: helper T ($0.34 \pm 0.15$, $0.98 \pm 0.01$, $0.03 \pm 0.01$), epithelial ($0.83 \pm 0.06$, $0.82 \pm 0.05$, $0.25 \pm 0.04$), progenitor ($0.47 \pm 0.1$, $0.98 \pm 0.01$, $0.07 \pm 0.02$), goblet ($0.91 \pm 0.01$, $0.97 \pm 0.0$, $0.19 \pm 0.03$), fibroblasts ($0.38 \pm 0.03$, $0.96 \pm 0.01$, $0.09 \pm 0.02$), and stromal ($0.74 \pm 0.03$, $0.74 \pm 0.03$, $0.29 \pm 0.03$). From these values, when considering prevalence, we note PPV is high for helper T and progenitor nuclei, and NPV is high for goblet and epithelial nuclei. A visualization of the virtual H&E performance on whole slide images is shown in Figure 5.

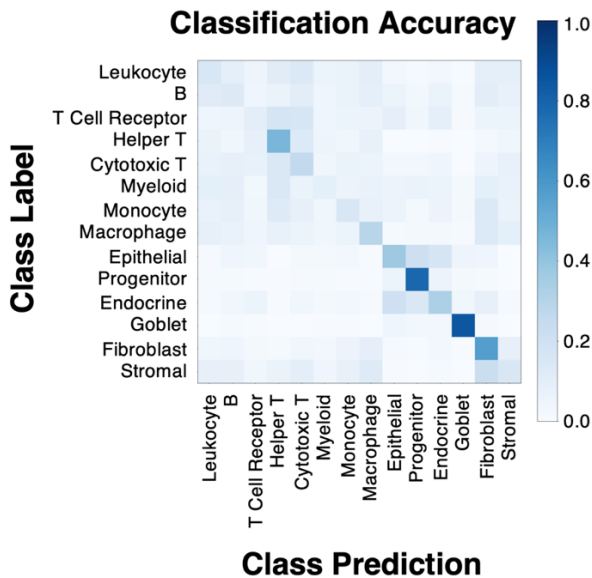

Figure 3. While not all classes were learned from virtual H&E, (shown for 5-fold cross-validation) some show learning. Learning behavior can be seen for helper T, macrophage, epithelial, progenitor, endocrine, goblet, and fibroblast nuclei. The model's ability to learn MxIF label information (derived from 17 stain channels) on virtual H&E (3 RGB channels) implies that there is signal present in our virtual H&E to learn some of our fine-grained nucleus subtypes.

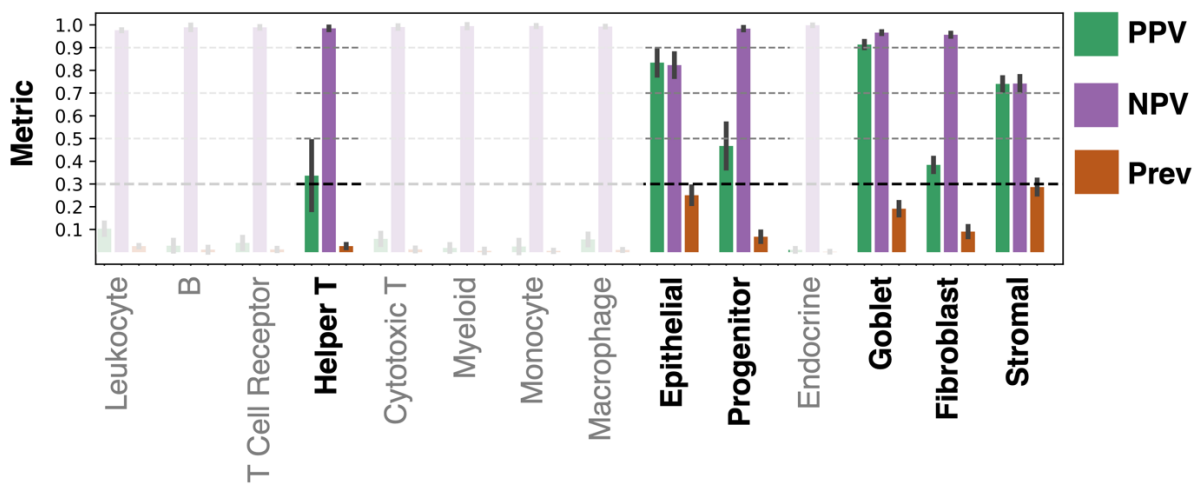

Figure 4. Learning MxIF stain information from virtual H&E is a challenging task, as illustrated by the positive predictive value (PPV), negative predictive value (NPV), and prevalence (Prev) for our pipeline across the 14 nucleus classes. The bar plot gives the mean value and each error bar denotes ± the standard deviation from cross-validation. We focus on the classes that could be identified with a mean PPV above 0.3, while the remaining classes are faded indicating lack of learning. The use of PPV, NPV, and prevalence allows us to better understand how the model would perform, if extended to unlabeled data, than the accuracies given in Figure 3.

## 4. CONCLUSIONS

In this work, we leveraged inter-modality learning to train models to classify 14 cell subtypes with virtual H&E input and MxIF-derived labels. The models were able to learn to identify nuclei on virtual H&E with PPV > 0.3 for these nucleus types: helper T (PPV > 0.3), progenitor (PPV > 0.45), epithelial (PPV > 0.8), goblet (PPV > 0.9), fibroblast (PPV > 0.35), and stromal (PPV > 0.7) (Figure 4).

While it is feasible to create a large number of labels for cell subtypes from MxIF, many of these labels are not easily learned on paired H&E-like data. This is not surprising, as specialized stains are commonly used to isolate many of the cells we attempted to learn to identify, such as helper T cells. However, for helper T, progenitor, epithelial, goblet, fibroblast and stromal nuclei, there is some learnable information in our virtual H&E, as evidenced by Figure 4. This signal may exist in real H&E, or it could be latent information that has resulted from the style transfer process from MxIF to virtual H&E.

Due to the lack of previous works performing multi-class nucleus classification on H&E using MxIF stain label information, it is difficult to compare the performance of our model to the literature. In the similar work from Han et al., 4 types of nuclei (ER+, PR+, HER2+, and Ki67+) were identified from real H&E information with AUCs >= 0.75 using MxIF stain label information[7]. These markers were used to assess breast cancer samples and were not present in our marker panel and their metric is for binary classification rather than multi-class classification, so a direct quantitative comparison is not reasonable.

Our work is limited by training and testing on virtual H&E, but not testing generalization on real H&E. Additionally, we only performed classification of nuclei using ground truth centroid information. Realistically, this type of model cannot be deployed on an unlabeled dataset because it only performs classification and does not locate nuclei. One might consider extending our proposed framework to include testing on real H&E data and incorporating a segmentation model to predict the locations of the nuclei to be classified.

Classification of nuclei on virtual H&E is promising for helper T, goblet, progenitor, epithelial, fibroblast, and stromal cells. Our approach is the first to identify helper T (PPV 0.34 ± 0.15; prevalence 0.03 ± 0.01) and progenitor cells (PPV 0.47 ± 0.1; prevalence 0.07 ± 0.02) on virtual H&E. The ability to discern nucleus subtypes based on shape and H&E staining is an exciting prospect in computational pathology.

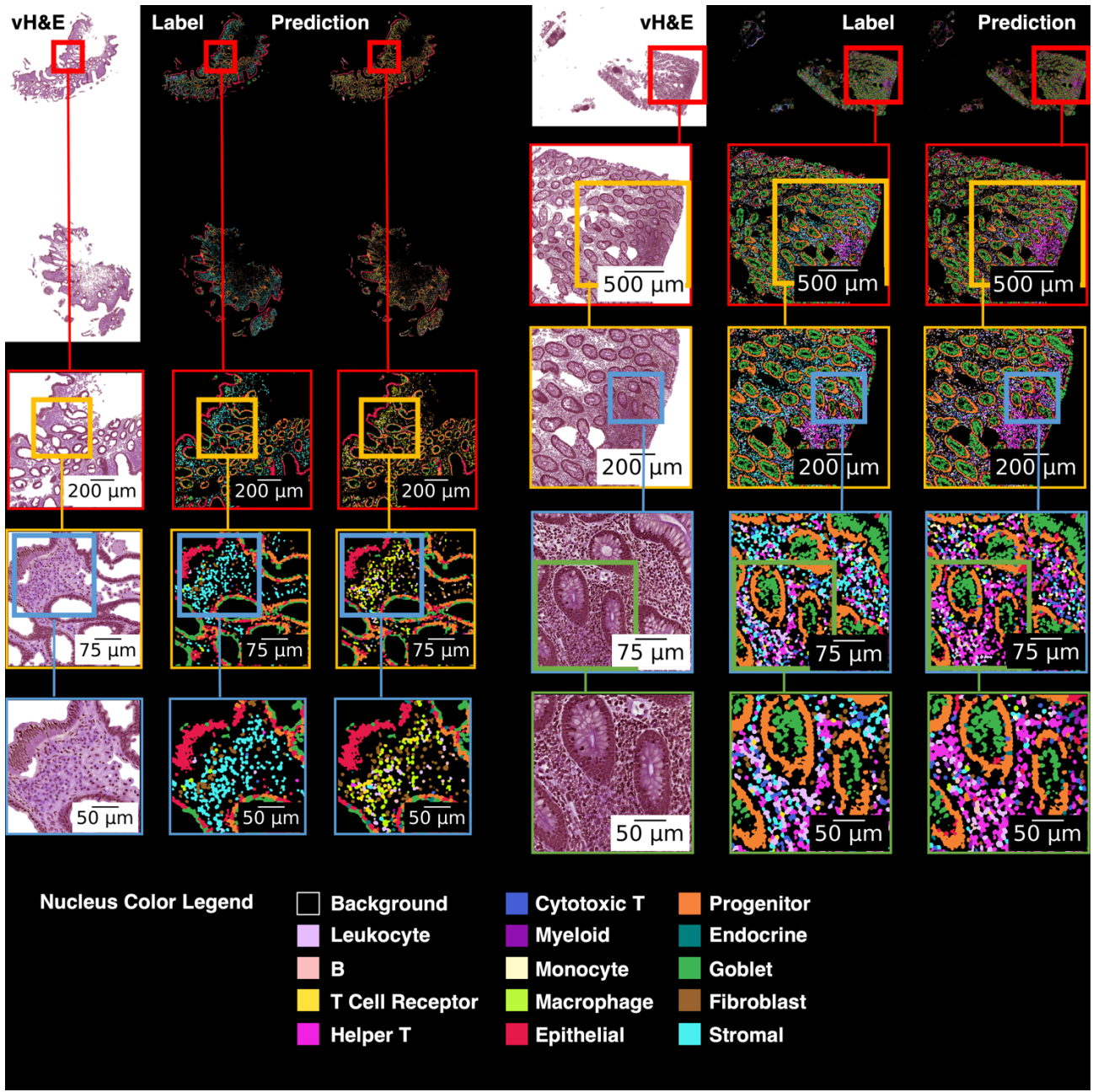

Figure 5. This figure shows virtual H&E, cell labels, and predicted labels for two whole slide images. At the top of the figure the full images are shown, and each subsequent row shows a magnified region of interest. On the left image, when fibroblasts are incorrectly predicted, the label is often stromal (both cell types are connective tissue cells). Stromal cells are often incorrectly classified as a variety of immune cells. On the right image, nuclei in an area with immune cell activity are often correctly identified as helper T cells, though the predictions do include a non-trivial number of false positives for helper T.

# ACKNOWLEDGMENTS

This publication is part of the Gut Cell Atlas Crohn's Disease Consortium funded by The Leona M. and Harry B. Helmsley Charitable Trust and is supported by a grant from Helmsley to Vanderbilt University www.helmsleytrust.org/gut-cell-atlas/. This research was supported by NSF CAREER 1452485, NSF 2040462, and in part using the resources of the Advanced Computing Center for Research and Education (ACCRE) at Vanderbilt University, Nashville, TN. This work was supported by Integrated Training in Engineering and Diabetes, grant number T32 DK101003. This material is partially supported by the National Science Foundation Graduate Research Fellowship under Grant No. DGE-1746891. This work was supported by the Department of Veterans Affairs I01BX004366, I01CX002171, and I01CX002473. We would like to acknowledge the VUMC Digestive Disease Research Center supported by NIH grant P30DK058404. This work is supported by NIH grant T32GM007347, R01DK135597, R01DK103831, R01DK128200. We extend gratitude to NVIDIA for their support by means of the NVIDIA hardware grant. This project was supported in part by the National Center for Research Resources, Grant UL1 RR024975-01, and is now at the National Center for Advancing Translational Sciences, Grant 2 UL1 TR000445-06. The Vanderbilt Institute for Clinical and Translational Research (VICTR) is funded by the National Center for Advancing Translational Sciences (NCATS) Clinical Translational Science Award (CTSA) Program, Award Number 5UL1TR002243-03. The content is solely the responsibility of the authors and does not necessarily represent the official views of the NIH.